\documentclass{article}
\usepackage{spconf,amsmath,graphicx}
\usepackage{enumitem}

\usepackage{graphicx}

\newcommand{\ie}{\textit{i}.\textit{e}.\space}

\usepackage{multirow,framed}
\usepackage{amsmath}
\usepackage{pifont}
\usepackage[normalem]{ulem}
\usepackage{amssymb}
\usepackage{latexsym}
\usepackage{url}
\usepackage{xcolor}
\usepackage{hyperref}
\usepackage{diagbox}
\usepackage{tabularx}
\usepackage{array}
\usepackage{ragged2e}
\usepackage{graphicx}
\usepackage{tikz}
\usepackage{bm}

\usepackage[utf8]{inputenc}
\usepackage{dashbox}
\newcommand\dboxed[1]{\dbox{\ensuremath{#1}}}
\renewcommand\fbox{\fcolorbox{red}{white}}

\usepackage{tabularx}

\setlist{nosep, leftmargin=14pt}

\usepackage{mwe} 

\title{Towards Classifying Histopathological Microscope Images as Time Series Data}
\name{Sungrae Hong$^{1}$ \space\space Hyeongmin Park$^{1}$ \space\space Youngsin Ko$^{2}$ \space\space Sol Lee$^{1}$ \space\space Bryan Wong$^{1}$ \space\space Mun Yi$^{1}$\sthanks{Corresponding Author}}

\address{$^{1}$Korea Advanced Institute of Science and Technology, Daejeon, South Korea
\\
    $^{2}$Seegene Medical Foundation, Seoul, South Korea}
\begin{document}
%
\maketitle
\begin{abstract}
As the frontline data for cancer diagnosis, microscopic pathology images are fundamental for providing patients with rapid and accurate treatment. However, despite their practical value, the deep learning community has largely overlooked their usage. This paper proposes a novel approach to classifying microscopy images as time series data, addressing the unique challenges posed by their manual acquisition and weakly labeled nature. The proposed method fits image sequences of varying lengths to a fixed-length target by leveraging Dynamic Time-series Warping (DTW). Attention-based pooling is employed to predict the class of the case simultaneously. We demonstrate the effectiveness of our approach by comparing performance with various baselines and showcasing the benefits of using various inference strategies in achieving stable and reliable results. Ablation studies further validate the contribution of each component. Our approach contributes to medical image analysis by not only embracing microscopic images but also lifting them to a trustworthy level of performance.

\end{abstract}

\section{Introduction}

Cancer remains a leading cause of death worldwide, prompting the deep-learning based computer vision community to develop various models to address this critical issue~\cite{echle2021deep}. Histopathology image classification models seek to assist medical professionals in reducing diagnostic errors and improving prognosis effectiveness~\cite{van2021deep}. However, we note that the deep learning community has neglected microscopy images, which have significant practical advantages over scanned images. Specifically, nearly every existing research has been conducted on scanner-based images. Microscopy images, a crucial tool for initial cancer diagnosis in clinical practice, offer the advantage of rapid results release~\cite{feng2020deep}. Moreover, microscopes are significantly more affordable than scanners~\cite{zhang2019whole} (\ie scanners cost tens of thousands of dollars on average, while microscopes cost only thousands), thereby offering significantly different accessibility to patients in third-world nations. Therefore, research based on microscopy images can open up new, improved opportunities for patients in those nations. 

\begin{figure}
 \centering\includegraphics[width=0.9\columnwidth]{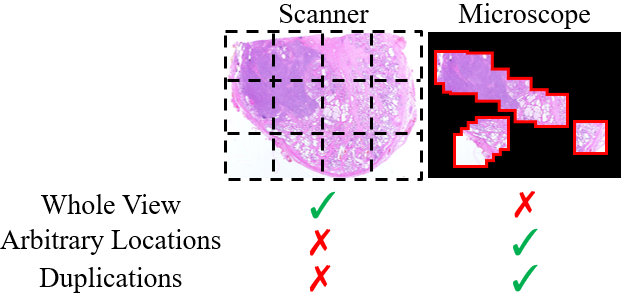}
 \caption[]{\dboxed{\phantom{1}} denotes an image patch acquired by the scanner and 
 \fbox{\phantom{1}} 
  a region captured by a microscope. 
 Microscopy images are composed of overlapping sequences manually captured by domain experts. 
 }
 \label{fig:1}
\end{figure}

\begin{figure*}[htbp]
\centering
\includegraphics[width=0.9\textwidth]{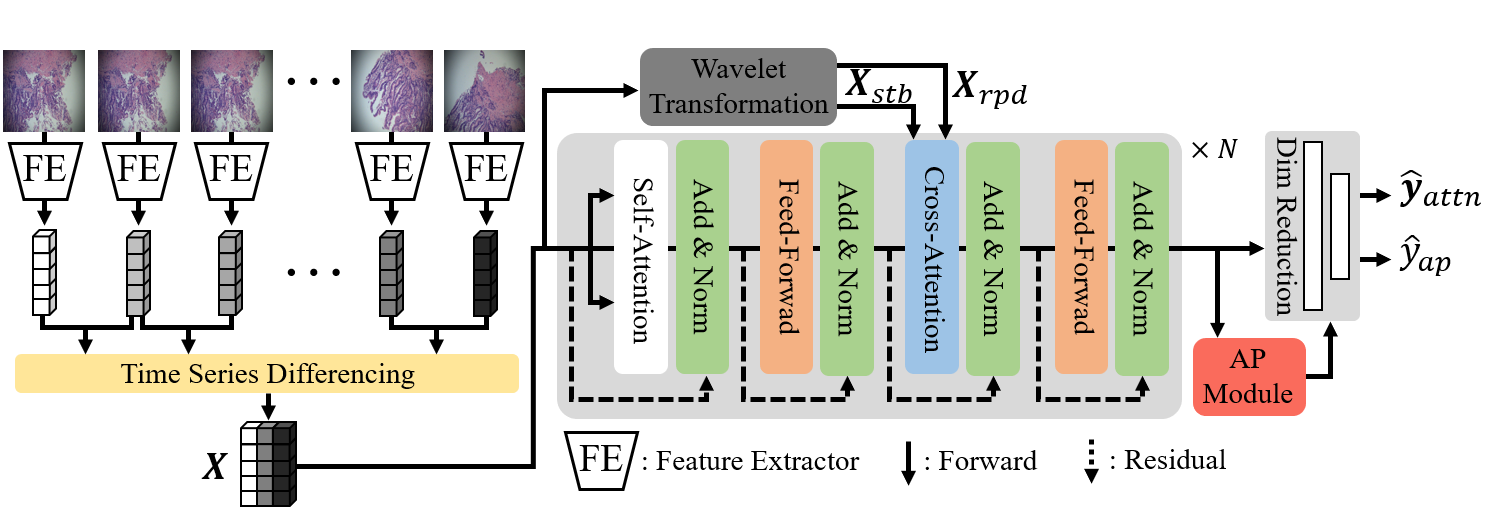}
\caption{Overall Framework}
\label{fig:1.overview}
\end{figure*}

Microscopy images present several challenges, making existing approaches for scanner-based research difficult to apply (\autoref{fig:1}). Among them are (1) \textit{ArbitraryLocations}: Microscopic images are periodically produced with a preset interval by a camera attached to the microscope while the microscope is being used by a pathologist or a medical doctor, thereby making it impossible to accurately capture the coordinates of where the produced images were acquired; (2) \textit{InconsistentLength}: the number of images captured per slide varies due to the automatic image acquisition process, which is easy to fool fixed-length approaches~\cite{sherstinsky2020fundamentals}; (3) \textit{Duplications}: the image  acquisition process often results in a large number of redundant images; and (4) \textit{WeakLabel}: the labels are weakly annotated. Whole-slide images (WSIs) from scanners allow pathologists to draw annotations on the entire tissue at a megapixel level~\cite{farahani2015whole}. However, the absence of a whole-view,  hundreds of images produced, and images of low-resolution make it very difficult for pathologists to annotate each image separately. Consequently, only a class label for an entire image set is available. To address these issues, in this paper we propose a novel approach specifically developed for microscopy histopathology images.

Although the microscopy images do not contain positional information, sequential acquisition over time enables us to treat them as image-time series data. Thus, we frame this problem as a time series classification task. 
To optimize inconstant length of input and its target, we leverage Dynamic Time-series Warping (DTW)~\cite{cuturi2017soft}. Redundant images are removed using time series differencing. If an image sequence is labeled with a specific class, it implies that at least one image within the sequence belongs to that class, regardless of the order of the images. Inspired by this property, we compress the temporal information of sequences using attention-based pooling. Additionally, we introduce a term that harmonizes time series prediction and compressed information prediction.

We summarize our contribution as follows:
\begin{itemize}
  \item{We address the unique challenges of microscopy data by formulating time series analysis. To the best of our knowledge, this is the first approach to tackle the constraints of microscopic images.}
  \item{We utilize a methodology, specifically designed to handle inconsistent sequence lengths, for classification. Additionally, we employ an Attention Pooling module to generate a summary of the entire image sequence.} 
  \item{Through various experiments, we demonstrate that our proposed approach provides a robust solution to the weakly labeled microscopy image classification problem.}
\end{itemize}

\section{Method}
We propose a method for classifying microscopic pathology image-sequences with varying lengths and weak labels. The overall framework is depicted in \autoref{fig:1.overview}.

\subsection{Data processing}
\begin{equation}\label{eq:1}
\Delta(\bm{x}_{i},\bm{x}_{i+1})=||{\bm{x}_{i+1}-\bm{x}_{i}}||_2^2
\end{equation}
\textbf{Time series differencing}
We extract a feature sequence $\textit{\textbf{X}}=\{\bm{x}_i\in\mathbb{R}^{d}{\,\mid\,}i=1,2,\cdots,n\}$, where $n$ is the length of the sequence and $d$ is feature dimension, from microscope images using a pre-trained feature extractor~\cite{kang2023benchmarking} on histopathology images. We applied time series differencing $\Delta(\cdot,\cdot)$ on every adjacent features $\bm{x}_i$ and $\bm{x}_{i+1}$ to remove redundant features with an L2 distance below the threshold $\tau$, determined via the validation set (\autoref{eq:1}).\\
\textbf{Wavelet transformation}
We investigate two key characteristics within image sequences: stable patterns and abrupt changes. Experts often capture narrow regions, causing similar symptoms to be filmed repeatedly, due to the manual zooming on the slide glass. On the other hand, they occasionally move the slide glass to acquire images from entirely different regions. Inspired by these distinct image acquisition patterns, we employ Wavelet transformation to extract low-frequency components $\textit{\textbf{X}}_{stb}\in\mathbb{R}^{n\times{d}}$ representing stable patterns and high-frequency components $\textit{\textbf{X}}_{rpd}\in\mathbb{R}^{n\times{d}}$ representing rapid changes.

\subsection{Attention module}
\begin{equation}\label{eq:attention}
    \text{Attention}(Q,K,V)=\text{sigmoid}\left(\frac{QK^T}{\sqrt{d_k}}\right){V}
\end{equation}
\begin{multline}\label{eq:crossattention}
    \text{CrossAttention}=    \lambda_{stb}\cdot\text{Attention}(Q,K_{stb},V_{stb})\\
    +\lambda_{rpd}\cdot\text{Attention}(Q,K_{rpd},V_{rpd})
\end{multline}

\begin{figure}[htb]\label{fig:seqtarget}
\begin{minipage}[b]{.48\linewidth}
  \centering
  \centerline{\includegraphics[width=5.0cm]{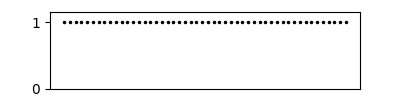}}
  \centerline{(a) Constant target}\medskip
\end{minipage}
\hfill
\begin{minipage}[b]{0.48\linewidth}
  \centering
  \centerline{\includegraphics[width=5.0cm]{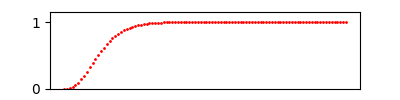}}
  \centerline{(b) Implicit target}\medskip
\end{minipage}
\caption{(a) and (b) is the sequence of the target class (\ie its shape is $\mathbb{R}^{{l}\times{1}}$). Unlike (a), (b) depicts a behavior of gradually observing regions corresponding to the class.}
\label{fig:res}
\end{figure}

\textbf{Self-Attention(SA)}
In scenarios where the target is weakly annotated, the model is encouraged to construct instance representations that capture the key features for classification. As shown in \autoref{eq:attention}, the Self-Attention module~\cite{vaswani2017attention} learns to organize representations by attending to relevant information within the sequence. Here, $Q$, $K$, and $V$ are defined as $\textbf{\textit{X}}W^Q$, $\textbf{\textit{X}}W^K$, and $\textbf{\textit{X}}W^V$, respectively, where $W^{Q,K,V}\in\mathbb{R}^{d\times{d_{k,q,v}}}$.\\
\textbf{Cross-Attention(CA)}
To enrich the representation of the input sequence, both $\textbf{\textit{X}}_{stb}
$ and $\textbf{\textit{X}}_{rpd}$ are integrated. We adopt a Cross-Attention module to selectively combine the informative components from the two inputs~\cite{ye2023ip} (\autoref{eq:crossattention}). The shapes of $Q$, $K$, $V$, and the transformation matrix $W$ used in CA are identical to those employed in SA. By assigning equal weights of $\lambda_{stb}=\lambda_{rpd}=1$, we aim for both stable and rapidly changing patterns to be equally considered.

\subsection{Attention Pooling(AP) module}
\begin{equation}\label{eq:AP}
    \textbf{\textit{P}}=\sum_{i=1}^{n}{a_i}{\bm{x}_i}\in\mathbb{R}^{d}
\end{equation}
\begin{equation}\label{eq:a_k}
    a_i=\frac{\exp{\{\bm{w}^T(\text{tanh}(\textit{\textbf{A}}_{t}\bm{x}_i)\,{\odot}\,\text{sigmoid}(\textit{\textbf{A}}_{s}\bm{x}_i))\}}}
    {\sum_{j=1}^{n}{\exp{\{\bm{w}^T(\text{tanh}(\textit{\textbf{A}}_{t}\bm{x}_j)\,{\odot}\,\text{sigmoid}(\textit{\textbf{A}}_{s}\bm{x}_j))\}}}}
\end{equation}
Because a sequence of length $n$ with a weak label indicates the presence of at least one relevant feature, the AP module aggregates the sequence into a $d$-dimensional embedding space using gated attention (\autoref{eq:AP} and \ref{eq:a_k}). In this context, $\bm{w}\in\mathbb{R}^{d}$ and $\textbf{\textit{A}}_{t},\textbf{\textit{A}}_{s}\in\mathbb{R}^{{d}\times{d}}$ are learnable parameters, where attention score value $a_i$ is a scalar.

\subsection{Identical dimension reduction network}
Although the Attention and AP modules predict different amounts of data (\ie$n$ and $1$), they share the common goal: mapping the representation to $C$ target classes. Furthermore, by jointly applying a dimension reduction network $f_\theta(\cdot)$ to single and weakly annotated sequence prediction, it can help to build richer model features for predicting $n$ multiple instances. Therefore, $f_\theta(\cdot)$ is shared across the modules. The $f_\theta(\cdot)$ consists of two linear networks with a ReLU activation in between, outputs $f_\theta(\text{CA}(\text{SA}(\textit{\textbf{X}}), \textit{\textbf{X}}_{stb}, \textit{\textbf{X}}_{rpd}))=\hat{\bm{y}}_{attn}\in\mathbb{R}^{n\times{C}}$ and $f_\theta(\textbf{\textit{P}})=\hat{y}_{ap}\in\mathbb{R}^C$.

\subsection{Objective functions}
\begin{equation}\label{eq:dtw}
    \mathcal{L}_{dtw}=||D(\hat{\bm{y}}_{attn},\bm{y}_l)-D(\bm{y}_{ideal},\bm{y}_l)||_1
\end{equation}
\begin{equation}\label{eq:align}
    \mathcal{L}_{align}=||\underset{C_i\in\{C_1,\cdots,{C_n}\}}{\operatorname{argmax}}(\hat{\bm{y}}_{attn})-\underset{C}{\operatorname{argmax}}(\hat{y}_{ap})||_1
\end{equation}
\begin{equation}
  \mathcal{L} = \lambda_{dtw}\mathcal{L}_{dtw} + \lambda_{ap}\mathcal{L}_{ap} + \lambda_{align}\mathcal{L}_{align}
\end{equation}
\\
\textbf{Soft-DTW term $\mathcal{L}_{dtw}$ and implicit time series target}
DTW algorithm~\cite{sakoe1971dynamic} measures the similarity between two sequences of varying durations. The algorithm maps dynamic-length sequences closer to pre-defined length of $l$ target sequence $\bm{y}_l\in\mathbb{R}^{l\times{C}}$. Because this is non-differentiable dynamic programming, we leverage the novel solution, Soft-DTW~\cite{cuturi2017soft} $D(\cdot,\cdot)$. Also, we define an ideal reference sequence $\bm{y}_{ideal}\in\mathbb{R}^{n\times{C}}$ to set the minimum value of the term to zero (\autoref{eq:dtw}). Meanwhile, we propose using the cumulative distribution function of a $\text{Beta}(3,20)$ as an implicit target to address the weakly supervised nature of our image sequences (\autoref{fig:res}). It is grounded in the empirical observation that experts find it challenging to capture regions exhibiting symptoms in a first frame due to the manual, high-zoom nature of the microscope.\\
\textbf{Attention-pooled target term $\mathcal{L}_{ap}$}
The AP module aggregates all the features in sequence into a prediction $\hat{y}_{ap}$. Therefore, $\mathcal{L}_{ap}$ is the cross-entropy loss between the weakly label and $\hat{y}_{ap}$.
\\
\textbf{Align term $\mathcal{L}_{align}$}
harmonizes the predicted class of $\hat{\bm{y}}_{attn}$ and $\hat{y}_{ap}$ (\autoref{eq:align}). Although $\textbf{\textit{X}}$ has a weak label, the most probable prediction in the sequence should be aligned with the aggregated embedding prediction.\\
\textbf{Total loss term $\mathcal{L}$} is weighted sum of $\mathcal{L}_{dtw}$, $\mathcal{L}_{ap}$ and $\mathcal{L}_{align}$, where $\lambda_{dtw},\lambda_{ap}$ and $\lambda_{align}$ are $1,10$ and $10$ respectively.
\section{Experiment}

\begin{table}[]
\centering
\caption{Dataset description. N and M indicates normal and malignant respectively.}
\label{tab:dataset}
\resizebox{\columnwidth}{!}{%
\begin{tabular}{c|c|cccccc}
\hline
\multirow{2}{*}{} &
  \multirow{2}{*}{Magnitudes} &
  \multicolumn{2}{c}{$n$(Images)} &
  \multicolumn{2}{c}{$n$(Cases)} &
  \multicolumn{2}{c}{\begin{tabular}[c]{@{}c@{}}Average\\ Images/Case\end{tabular}} \\ \cline{3-8} 
                                                                      &                   & N   & M     & N                   & M                   & N  & M  \\ \hline
\multirow{4}{*}{BreakHis}                                             & $40\times$               & 652 & 1,370 & \multirow{4}{*}{24} & \multirow{4}{*}{58} & 27 & 24 \\
                                                                      & $100\times$              & 644 & 1,437 &                     &                     & 27 & 25 \\
                                                                      & $200\times$              & 623 & 1,390 &                     &                     & 26 & 24 \\
                                                                      & $400\times$              & 588 & 1,232 &                     &                     & 15 & 21 \\ \hline
{\begin{tabular}[c]{@{}c@{}}SMF\end{tabular}} & $40\times, 100\times, 200\times$ & 52,076   & 84,100     & 492                   & 423                   & 106  & 199  \\ \hline
\end{tabular}%
}
\end{table}

\begin{table*}[]
\centering
\caption{Comparison results}
\label{tab:quantitative}
\resizebox{1.8\columnwidth}{!}{%
\begin{tabular}{cc|cc|cccccccc}
\hline
\multicolumn{2}{l|}{\multirow{3}{*}{}} &
  \multicolumn{2}{c|}{\multirow{2}{*}{{\begin{tabular}[c]{@{}c@{}}SMF\\ (n=186)\end{tabular}}}} &
  \multicolumn{8}{c}{BreakHis \textit{(n=28)}} \\ \cline{5-12} 
\multicolumn{2}{l|}{} &
  \multicolumn{2}{c|}{} &
  \multicolumn{2}{c}{40$\times$} &
  \multicolumn{2}{c}{100$\times$} &
  \multicolumn{2}{c}{200$\times$} &
  \multicolumn{2}{c}{400$\times$} \\ \cline{3-12} 
\multicolumn{2}{l|}{} &
  F1 &
  Accuracy &
  F1 &
  Accuracy &
  F1 &
  Accuracy &
  F1 &
  Accuracy &
  F1 &
  Accuracy \\ \hline
\multicolumn{2}{c|}{LSTM~\cite{hochreiter1997long}} &
  0.975 &
  0.976 &
  0.940 &
  0.914 &
  0.923 &
  0.893 &
  0.938 &
  0.914 &
  0.925 &
  0.893 \\
\multicolumn{2}{c|}{GRU~\cite{chung2014empirical}} &
  0.978 &
  0.980 &
  0.916 &
  0.879 &
  0.945 &
  0.921 &
  0.933 &
  0.907 &
  0.927 &
  0.900 \\
\multicolumn{2}{c|}{Transformer~\cite{vaswani2017attention}} &
  0.981 &
  0.983 &
  0.928 &
  0.900 &
  0.943 &
  0.921 &
  0.932 &
  0.907 &
  0.949 &
  0.929 \\
\multicolumn{2}{c|}{ABMIL~\cite{ilse2018attention}} &
  0.982 &
  0.984 &
  0.891 &
  0.850 &
  0.913 &
  0.871 &
  0.911 &
  0.871 &
  0.904 &
  0.857 \\
\multicolumn{2}{c|}{TransMIL~\cite{shao2021transmil}} &
  0.977 &
  0.978 &
  0.941 &
  0.914 &
  0.924 &
  0.893 &
  0.918 &
  0.886 &
  0.924 &
  0.893 \\ \hline
\multicolumn{1}{c|}{\multirow{4}{*}{Ours}} &
  AP &
  \textbf{0.990} &
  \textbf{0.991} &
  0.954 &
  0.936 &
  0.947 &
  0.929 &
  0.945 &
  0.921 &
  0.954 &
  0.936 \\ \cline{2-12} 
\multicolumn{1}{c|}{} &
  DTW Distance &
  0.988 &
  0.989 &
  0.962 &
  \textbf{0.950} &
  0.951 &
  \textbf{0.936} &
  0.947 &
  0.929 &
  0.969 &
  \textbf{0.957} \\
\multicolumn{1}{c|}{} &
  KNN &
  \textbf{0.990} &
  \textbf{0.991} &
  0.954 &
  0.936 &
  \textbf{0.952} &
  \textbf{0.936} &
  0.953 &
  0.936 &
  \textbf{0.970} &
  \textbf{0.957} \\ \cline{2-12} 
\multicolumn{1}{c|}{} &
  Voting &
  \textbf{0.990} &
  \textbf{0.991} &
  \textbf{0.964} &
  \textbf{0.950} &
  \textbf{0.952} &
  \textbf{0.936} &
  \textbf{0.958} &
  \textbf{0.943} &
  0.964 &
  0.950 \\ \hline
\end{tabular}%
}
\end{table*}

\textbf{Experimental settings}
We adopted the \textit{Tiny} recipes from \cite{dosovitskiy2020image} for the attention module's parameters. The model was trained using the Adam optimizer~\cite{kingma2014adam} with a learning rate of $1e-4$ and betas of $(0.9, 0.999)$. All experiments were conducted  on a single $\text{NVIDIA}^\circledR$ RTX 2080 Ti.\\
\textbf{Dataset}
We employ two microscopy datasets: BreakHis~\cite{spanhol2015dataset}, a benchmark of breast histopathology images with separated magnifications, and private Seegene Medical Foundation (SMF), colon polyp images with mixed magnifications.
While BreakHis offers pre-split train/test sets, SMF did not. Thus, we randomly selected 20\% of the entire SMF as the test set. For both datasets, 15\% of the training data was randomly selected as a validation set. We set differencing threshold of $\tau$. For SMF, it was set to tackle 25\% of duplicate images. For BreakHis, a lower $\tau$ was selected so as 
 to eliminate 5\%  of them, due to its relatively fewer redundancies. For BreakHis, $\bm{y}_{25}$, the average number of images per case, was used; for SMF, $\bm{y}_{75}$, which is smaller than its average length, was used for computational efficiency.\\
\textbf{Baselines}
We compare three baselines~\cite{hochreiter1997long,chung2014empirical,vaswani2017attention} that classify the entire time series as a single category. Additionally, we employ two Multiple Instance Learning (MIL) models~\cite{ilse2018attention,shao2021transmil} suitable for weak supervision. MIL classifies bags formed by randomly shuffling the set of images $\textbf{\textit{X}}$.

\subsection{Analysis on inference strategies}




 Our proposed method offers four inference strategies: $\hat{y}_{ap}$ prediction, DTW distance between $\hat{\bm{y}}_{attn}$ and $\bm{y}_l$, K-Nearest Neighbors (KNN) using $\hat{\bm{y}}_{attn}^{train}$ and $\hat{\bm{y}}_{attn}$, and a majority voting of them. 
 The performances of these strategies are quantified in \autoref{tab:quantitative} under the label "Ours". Majority voting generally yielded the best performance. Additionally, it always outperformed the inference strategy with the lowest performance among the three strategies. It indicates that, despite fluctuations in the individual inference strategies, the $\hat{y}_{ap}$ and $\hat{\bm{y}}_{attn}$ are complementary.

\begin{figure}
 \centering\includegraphics[width=0.85\columnwidth]{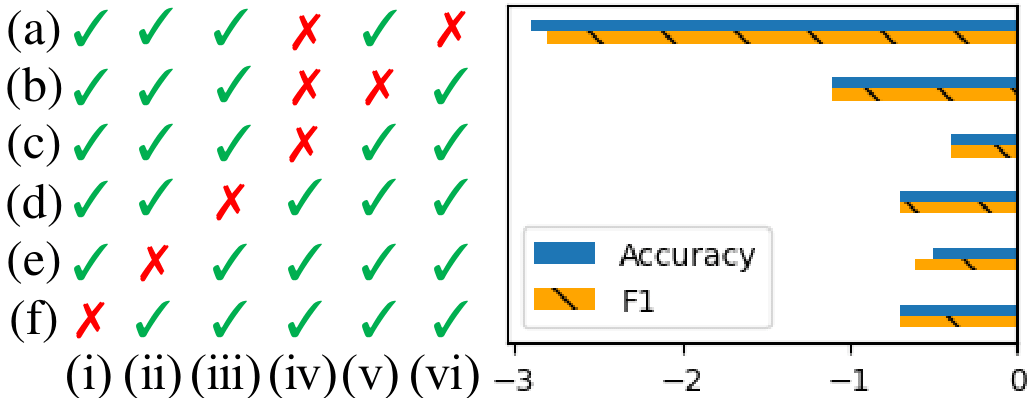}
 \caption[]{We ablate $(\mathrm{i})$ Wavelet transformation $(\mathrm{ii})$ Implicit target $(\mathrm{iii})$ Reference sequence $\bm{y}_{ideal}$ $(\mathrm{iv})$ $\mathcal{L}_{align}$ $(\mathrm{v})$ Attention module $(\mathrm{vi})$ AP module. We present combinations of ablations from $(\text{a})$ to $(\text{f})$ and plot the performance decreases compared to the fully equipped model.}
 \label{fig:ablation}
\end{figure}

\subsection{Results}
\textbf{Quantitative results}
\autoref{tab:quantitative} presents the performance comparisons between the proposed methodology and various baseline models. Our proposed method outperformed other models on the SMF dataset. Transformer and MIL-based models generally exhibited better performance than traditional time-series models. These findings indicate that more advanced models are advantageous for noisy and poorly curated data. We conducted experiments on four magnitudes of the BreakHis and our proposed method achieved superior results compared to other methods. In this dataset, sequence embedding-based methods outperformed MIL approaches. Our proposed method can predict the entire sequence without truncation and aggregate the entire sequence for a single prediction, combining the advantages of both approaches.
\\
\textbf{Ablation study}
We observed performance variations on the SMF dataset by ablating different components. Performance was measured using voting and plotted in percentage points. For experiments where $(\mathrm{v})$ and $(\mathrm{vi})$ were removed, $\mathcal{L}_{align}$ could not be measured, thus we also removed $(\mathrm{iv})$. Because $\hat{\bm{y}}_{attn}$ and $\hat{y}_{ap}$ is absent in (v) and (vi) respectively, voting could not be applied. Therefore, we used $\hat{{y}}_{ap}$ for $(\mathrm{v})$ and distance using $\hat{\bm{y}}_{attn}$ for $(\mathrm{vi})$ as prediction strategies. \autoref{fig:ablation} shows that the ablations resulted in an overall performance reduction. Notably, either sequence removal or single-point prediction removal resulted in the most substantial performance degradation ($(\text{a})$ and $(\text{b})$), which aligns with our previous findings indicating a complementary relationship between these two prediction methods.
\section{Conclusion}
Focusing on microscope images, a practically significant but under-explored type of visual data, this study formulated automatically captured image sequences as time series data. Our proposed framework preprocesses data using various techniques and performs time series analysis and point estimation simultaneously. Through extensive experiments, we have demonstrated that employing diverse inference strategies and combining their results through voting can yield superior outcomes. Our work expands the realm of medical image
analysis by embracing microscopic images and showcasing their usefulness via the proposed method, overcoming the unique challenges associated with the microscopy images.

 
\section{COMPLIANCE WITH ETHICAL STANDARDS}
This study was performed in line with the principles of the Declaration of Helsinki. Approval was granted by the Ethics Review Board (SMF-IRB-2020-007) and (KAIST-IRB-22-335). Also, this research study was conducted retrospectively using human subject data made available in open access by \cite{spanhol2015dataset}. Ethical approval was not required as confirmed by the license attached with the open access data.

\section{Acknowledgement}
This research was supported by the Seegene Medical Foundation, South Korea, under the project “Research on Developing a Next Generation Medical Diagnosis System Using Deep Learning” (Grant Number: G01180115).

\end{document}